\documentclass[sigconf]{aamas} 

\renewcommand\footnotetextcopyrightpermission[1]{} 
\usepackage{balance}

\usepackage[utf8]{inputenc}
\usepackage[ruled]{algorithm2e}
\usepackage{amsmath}
\usepackage{amsthm}
\usepackage{mathtools}
\mathtoolsset{showonlyrefs=true}
\usepackage{hyperref}
\usepackage{makecell}
\usepackage{subcaption}
\usepackage{wrapfig}
\usepackage{etoolbox}

\title[AAMAS-2021 Formatting Instructions]{Self-Imitation Advantage Learning}

\author{Johan Ferret}
\affiliation{
  \institution{Google Research, Brain team}}
\affiliation{
  \institution{Inria Lille Nord Europe, SCOOL team}}
\email{jferret@google.com}

\author{Olivier Pietquin}
\affiliation{
  \institution{Google Research, Brain team}}
\email{pietquin@google.com}

\author{Matthieu Geist}
\affiliation{
  \institution{Google Research, Brain team}}
\email{mfgeist@google.com}

\begin{abstract}
    Self-imitation learning is a Reinforcement Learning (RL) method that encourages actions whose returns were higher than expected, which helps in hard exploration and sparse reward problems. It was shown to improve the performance of on-policy actor-critic methods in several discrete control tasks. Nevertheless, applying self-imitation to the mostly action-value based off-policy RL methods is not straightforward. We propose SAIL, a novel generalization of self-imitation learning for off-policy RL, based on a modification of the Bellman optimality operator that we connect to Advantage Learning. Crucially, our method mitigates the problem of stale returns by choosing the most optimistic return estimate between the observed return and the current action-value for self-imitation. We demonstrate the empirical effectiveness of SAIL on the Arcade Learning Environment, with a focus on hard exploration games.
\end{abstract}

\definecolor{citecolor}{rgb}{0,0.08,0.45}
\hypersetup{colorlinks,
linkcolor=citecolor,
citecolor=citecolor,
urlcolor=citecolor
}

\makeatletter
\patchcmd{\maketitle}{\@copyrightspace}{}{}{}
\makeatother

\makeatletter
\def\@copyrightspace{\relax}
\makeatother

\makeatletter
\renewcommand{\@algocf@capt@plain}{above}
\makeatother

\newcommand{\jf}[1]{{\color{blue}[jferret: #1]}}
\newcommand{\mfg}[1]{{\color{red}[mfgeist: #1]}}


\begin{document}


\pagestyle{fancy}
\fancyhead{}


\maketitle 

\section{Introduction}


Some approaches combine Reinforcement Learning (RL) and learning from (expert) demonstrations~\cite{piot2014boosted, hester2017}.
It is efficient, but having access to expert demonstrations might not be possible in many situations. An interesting and more practical alternative is to learn from oneself, by focusing on one's own instructive experiences. In the context of RL~\citep{sutton2018}, self-imitation learning is a technique that takes advantage of this idea and proposes to learn from positive experiences, using actions whose payoff was superior to what had been predicted. It was introduced by~\citet{oh2018} for on-policy learning, where interaction data is obtained via a behaviour policy and is used to improve this policy, never to be reused again. In that context, self-imitation provides a practical way to revisit and reinforce interesting actions.  

In the off-policy setting, agents learn instead from the behaviour of different policies than their own, while they are still periodically allowed to interact in the environment to generate new interaction data. Off-policy learning is attractive because it would make RL useful in situations where interacting in the environment is costly or impractical, which is the case in many real-world scenarios~\citep{dulac2019}. In the most extreme case, offline RL~\citep{levine2020}, no interaction at all is possible and one must learn a policy just by looking at a dataset of past experiences generated by potentially many different policies. Sadly, the canonical form of self-imitation learning relies on a modified policy gradient~\citep{sutton1984}, which requires the ability to modify the current policy in a given direction. This requirement makes it incompatible with most off-policy methods~\citep{mnih2013, hessel2017, kapturowski2018}, whose policy is obtained by applying an exploration method on top of the current estimate of the optimal action-value. Since the policy is implicit, gradient ascent in the policy space is impractical. As an example, a straightforward adaptation of SIL for a standard off-policy RL algorithm such as Deep Q-Networks~\citep[DQN,][]{mnih2013} seems out-of-reach. Indeed, the action-value network that is central to DQN encapsulates the two aspects SIL targets separately in the actor-critic: value estimation and decision-making. 

The focus of this work is on providing a self-imitation method that is applicable across the full spectrum of off-policy methods. A common denominator to all off-policy algorithms is the use of the action-value to inform decision-making, be it with an implicit policy
~\citep{mnih2013}, or an explicit one~\citep{haarnoja2018}. Hence, a way to control which actions are reinforced in off-policy learning is to artificially increase the reward of such actions. Staying in the line of reasoning of SIL, a natural idea is to use the difference between the observed return and the estimated value as the reward bonus. It turns out that doing so creates a problem because the observed return becomes stale over time, biasing action-value towards pessimism instead of optimism, and eventually reducing the self-imitation contribution to none. To circumvent this issue, we opt for the simple strategy of using the most optimistic between the return and the estimated action-value, with which we extend the benefits of self-imitation. We show a connection to Advantage Learning~\citep[AL,][]{baird1999, bellemare2016}, an action-gap increasing off-policy algorithm.

Our contributions are the following: 1) we propose SAIL, a generalization of self-imitation learning for off-policy methods, 2) we show how it connects to Advantage Learning and complements it, 3) we demonstrate the practicality of our method in terms of simplicity, efficiency and performance on the Arcade Learning Environment~\citep[ALE,][]{bellemare2013} benchmark, under several base off-policy RL methods. Notably, we report considerable performance gains on hard exploration games. 

\section{Background and Notations}

\subsection{Reinforcement Learning}

We use the standard Markov Decision Processes (MDP) formalism~\citep{puterman1994}. An MDP is a tuple $M = \{ \mathcal{S}, \mathcal{A}, P, R, \gamma \}$, where $\mathcal{S}$ is the state space, $\mathcal{A}$ is the action space, $P$ is the transition kernel, $R$ is the bounded reward function and $\gamma \in [0, 1)$ is the discount factor. 
We note $\tau = \{ s_i, a_i, r_i \}_{i = [0, T]}$ a (random) trajectory and the associated return is
$G = \sum_{t = 0}^{\infty} \gamma^t r_t$.
A policy $\pi \in \Delta_{\mathcal{A}}^\mathcal{S}$ outputs action probabilities in a given state, and can be used for decision-making in a given environment. We place ourselves in the infinite-horizon setting, which means that we look for a policy that maximizes $J(\pi) = \mathbb{E}_{\pi}[G_0]$.
The value of a state is the quantity $V_{\pi}(s) = \mathbb{E}_{\pi}[\sum_{t = 0}^{\infty} \gamma^t r_t | S_0 = s]$. The action-value of a state-action couple is the quantity $Q_{\pi}(s, a) = \mathbb{E}_{\pi}[\sum_{t = 0}^{\infty} \gamma^t r_t | S_0 = s, A_0 = a]$.
The Bellman operator $\mathcal{T}_\pi$ is
$ \mathcal{T}_\pi V(s) = \mathbb{E}_{\pi}[r(s, a) + \gamma V(s')]$, its unique fixed point is $V_\pi$.
The Bellman optimality operator $\mathcal{T^*}$ is defined as
  $\mathcal{T^*}Q(s, a) = \mathbb{E}_{\pi}[r(s, a) + \gamma \max_{a'} Q(s', a')]$, its unique fixed point is the optimal action value $Q^*$.
A policy belongs to the set of optimal policies iff $Q_{\pi}(s, a) = Q^*(s, a)$.
The action-gap~\cite{farahmand2011} $g_{Q^*}$ of a particular state quantifies the difference between the value of the best action and that of the other actions:
$ g_{Q^*}(s, a) = \max_{a'} Q^*(s, a') - Q^*(s, a)$.
The advantage $A$ is the difference between the action-value and the value:
$ A(s, a) = Q(s, a) - V(s)$.

\paragraph{Actor-critic}
Actor-critic methods use a policy and a value function. In the standard formulation, the value (from the critic) guides the policy via the policy gradient,
$\mathcal{L}_{PG} = - \log \pi_{\theta}(a_t | s_t) (G_t - V_{\theta}(s_t))$,
while the value itself is updated via temporal difference on data generated by the actor:
$\mathcal{L}_{V} = \frac{1}{2} (r_t + \gamma V_{\theta}(s_{t+1}) - V_{\theta}(s_t))^2$.

\paragraph{Off-policy RL}
DQN~\citep{mnih2015} is an adaptation of Q-learning~\citep{watkins1992} to the deep learning framework. It conserves the principal aspect of Q-learning, that is updating the current action-value towards a bootstrapped version of itself, following the temporal difference algorithm. The major difference with respect to Q-learning is the use of a function approximator for the action-value, instead of using a tabular representation. Past transitions are sampled from a replay buffer~\citep{lin1992}, and, for stability, the action-value to be updated is compared to a target function, which is a previous and periodically updated version of the action-value function (whose parameters we note $\theta
^-$). Rainbow~\citep{hessel2017} extends DQN by combining several independently proposed algorithmic innovations~\citep{schaul2015, van2015, wang2015, bellemare2017, fortunato2017}, and showed to be a strong baseline on the Atari Learning Environment benchmark. IQN~\citep{dabney2018} is a distributional RL approach that estimates the whole distribution of the returns, instead of the mean as in standard Q-learning. More precisely, it approximates the continuous quantile function of the return distribution by sampling input probabilities uniformly. In terms of performance, it (almost) bridges the gap with Rainbow while not having any prioritized replay and using single-step bootstrapping.


\subsection{Self-Imitation Learning}

Self-imitation learning~\citep[SIL,][]{oh2018} provides a set of additional loss functions that complement the existing actor-critic losses, and encourage the agent to mimic rewarding past behavior.  It relies on a replay buffer that stores past transitions $\{ s_t, a_t, G_t \}$, where the observed return $G_t = \sum_{i = 0}^{+\infty} \gamma^i r_{t + i}$ is obtained once the episode is over. SIL operates off-policy, since its losses are calculated using past transitions from the replay buffer, implying that on-policy actor-critic methods that benefit from SIL become part on-policy, part off-policy. 
For a given transition $(s_t, a_t, G_t)$, the additional policy and value losses are:
\begin{equation}
    \mathcal{L}^\text{sil}_{PG} = - \log \pi_{\theta}(a_t | s_t) (G_t - V_{\theta}(s_t))_+
    \text{ and }
    \mathcal{L}^\text{sil}_{V} = \frac{1}{2} (G_t - V_{\theta}(s_t))_+^2,
\end{equation}
%
where the notation $(x)_+ = \max(0, x)$ stands for the ReLU operator. 
The shared term $(G_t - V_{\theta}(s_t))_+$ is positive if the observed return outweighs the estimated value, in which case both the action probabilities and the value are increased.
SIL also relies on prioritization~\citep{schaul2015}, as transitions are sampled from the replay buffer with probability $P(\tau) \propto (G_t - V_{\theta^-}(s_t))_+$.  As a result, transitions in which actions led to an unexpectedly high return get revisited more often, ensuring faster propagation of important reward signals.

A drawback of this formulation is that an explicit policy is required.
 Also, while it provides good empirical results, self-imitation in general might suffer from stochasticity, both in the environment dynamics and in the variance of the rewards. Indeed, the optimistic updates of SIL can lead to overestimating the value of actions that bring high rewards once in a while. See Sec.~\ref{subsection:stochasticity} for a more in-depth discussion about the role of stochasticity.


\subsection{Advantage Learning}

Advantage Learning (AL) is an action-gap increasing algorithm introduced by~\citet{baird1999} and further studied by~\citet{bellemare2016}. 
It corresponds to the following modified Bellman operator,
$ \mathcal{T}_{AL}Q(s, a) = \mathcal{T}^*Q(s, a) + \alpha (Q(s, a) - \max_{a'} Q(s, a'))$,
which translates to the following reward modification:
\begin{equation*}
  \tilde{r}_{AL}(s, a) = r(s, a) + \alpha (Q(s, a) - \max_{a'} Q(s, a')).
\end{equation*}
It can be seen as adding the advantage $Q(s,a)-V(s)$ to the reward, with $V(s) = \max_a Q(s,a)$ (hence the name).
%
Generally speaking, increasing the action-gap makes the RL problem easier, as it facilitates the distinction from experience between the optimal action and the others.
%
In practice, AL was shown to bring consistent performance improvement across the whole support of Atari games for off-policy agents~\citep{bellemare2016}.


\section{SAIL: Self-Imitation Advantage Learning}
\label{section:sail}

In essence, self-imitation drives the policy towards actions whose returns are unexpectedly good. Off-policy methods usually optimize an action-value function, which only implicitly defines the policy through a given exploration method. Nevertheless, there is a sensible proxy for increasing the probability of picking an action: increasing its action-value (which is nothing more than a reward increase). Hence, we propose to adapt self-imitation for off-policy using the following modified reward:
$$\tilde{r}_{SAIL}(s_t, a_t) = r(s_t, a_t) + \alpha (\max(G_t, Q_{\theta^-}(s_t, a_t)) - \max_a Q_{\theta^-}(s_t, a)),$$
The corresponding loss function is:
\begin{equation}
\label{eqn:loss}
    \mathcal{L}_{SAIL} = \frac{1}{2} \left({\tilde{r}_{SAIL}(s_t, a_t) + \gamma \max_a Q_{\theta^-}(s_{t + 1}, a) - Q(s_t, a_t)}\right)^2.
\end{equation}
We now motivate this expression in more details. 

\begin{figure}
    \centering
    \includegraphics[width=0.57\linewidth]{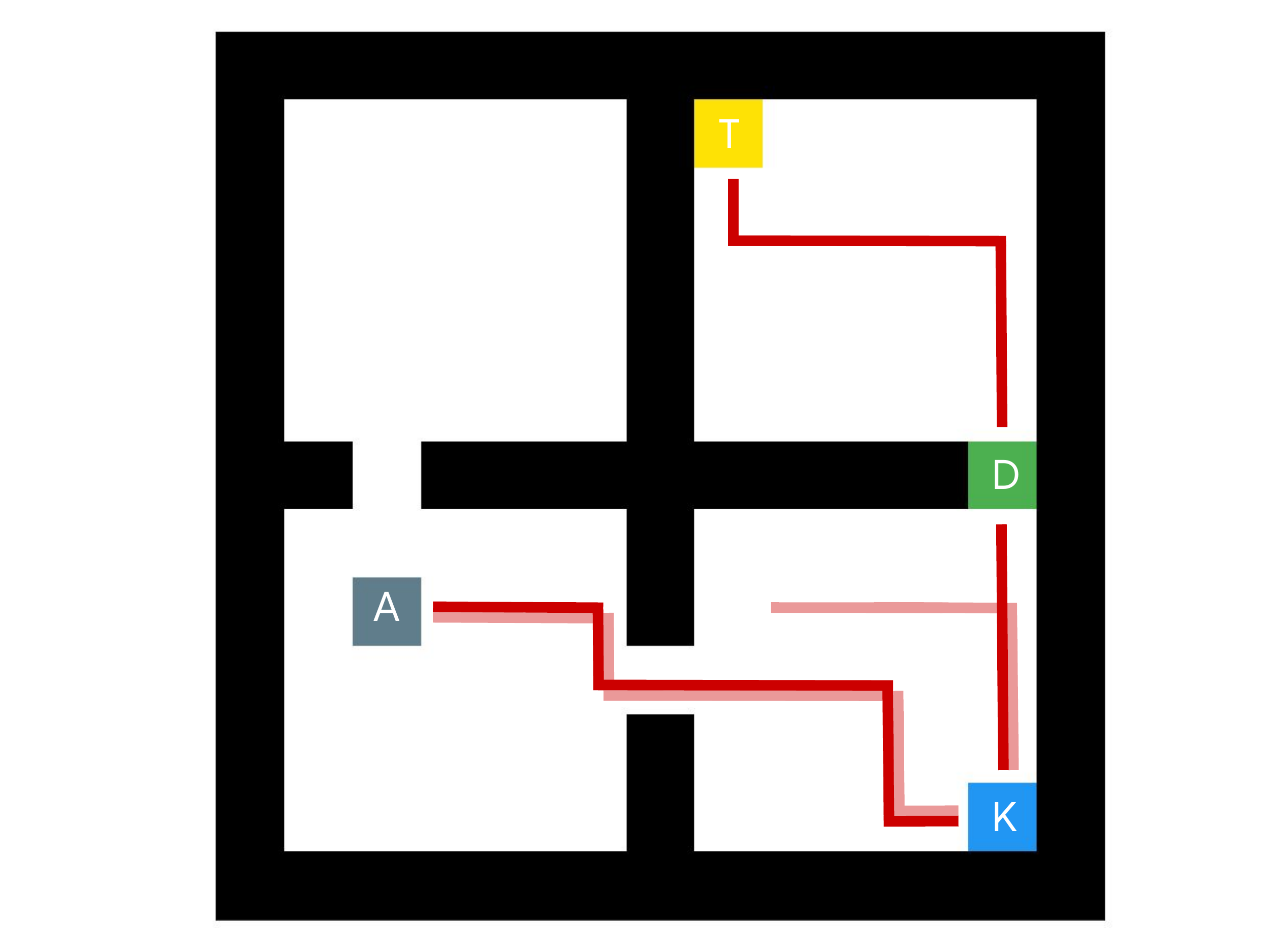}
    \caption{Illustration of the stale returns problem, in the Key-Door-Treasure environment~\citep{oh2018}. In past experiences (in faded red), the agent ("A") got the key ("K") but failed to reach the goal ("T"). The corresponding transitions still get sampled from the replay buffer, so actions leading to the key are paired with zero returns. This is inconsistent with the current policy (in red), which gets the key, opens the door and reaches the goal. SAIL avoids undervaluing the role of the key by using the most optimistic return estimate between the observed return and the current action-value.}
    \label{fig:kdt}
\end{figure}

First, note that if we replace $\max(G_t, Q(s, a))$ by $G_t$ and apply the ReLU operator to the additional term we get the modified reward:
\begin{equation}
\label{eqn:sil-loss}
\tilde{r}(s_t, a_t) = r(s_t, a_t) + \alpha (G_t - \max_a Q_{\theta^-}(s_t, a))_+,
\end{equation}
which would be a more straightforward adaptation of SIL, since it increases the action-value by the same term used in SIL for both policy and value updates. 
Empirically, this formulation compares unfavorably to ours (see Sec.~\ref{subsection:sil-vs-sail}). We posit it could be due to a limitation of self-imitation: the problem of stale returns. This problem is the following: when off-policy, the return used for self-imitation corresponds to the observed return of a policy that becomes increasingly different from the current one, under the hypothesis that the agent is still learning. Inevitably, returns become outdated. More precisely, these returns drift towards pessimism, since the agent mostly improves over time. As a result, the self-imitation bonus disappears for most transitions. In addition to making self-imitation ineffective, it could lead to emphasizing wrong signals and biasing the updates. We illustrate this phenomenon in Fig.~\ref{fig:kdt}. By exchanging the return for the maximum between the return and the current action-value estimate, our method forms a more optimistic return estimate, which promotes optimism and alleviates staleness. Additionally, by replacing the ReLU operator with the identity, we update our action-value estimate using information from both positive and negative experiences. Finally, compared to SIL, we do not make use of prioritization, as we think it emphasizes the impact of stochasticity on the algorithm. In our experiments, we measure the benefits of our method under increasing stochasticity in Sec.~\ref{subsection:stochasticity}, and show that the gains over the baseline remain even when stochasticity is high.
Second, the proposed modified reward can be decomposed as:
\begin{align*}
\tilde{r}_{SAIL}(s_t, a_t) 
&= \tilde{r}_{AL}(s_t, a_t) + \alpha (G_t - Q_{\theta^-}(s_t, a_t))_+,
\end{align*}
Thus, our approach actually combines AL, an action-gap increasing method, with SIL, which helps reproducing interesting trajectories from the past. We show next that it outperforms AL.

We name the resulting algorithm Self-Imitation Advantage Learning (SAIL\footnote{We take the liberty of inverting the middle letters to form a nice acronym.}). It is: \textbf{1) general}, i.e.\xspace compatible across a large spectrum of off-policy methods,
\textbf{2) easy to use}, as it arranges existing quantities to form a reward bonus, and \textbf{3) lightweight}, in the sense that it does not add much to the computational budget of the algorithm (other than computing discounted returns, which is neglectible). The pseudo-code is in Alg.~\ref{alg:1}.
\begin{algorithm}
\SetAlgoLined
\caption{SAIL: Self-Imitation Advantage Learning}
 Initialize the agent weights $\theta$\;
 Initialize the replay buffer $\mathcal{B}$\;
 Initialize the return placeholder $G_{\varnothing}$\;
 \For{each iteration}{
  \tcc{Collect and store interaction data.}
  \For{each interaction step}{
    In $s_t$, sample $a_t \sim \pi_{\theta}$, act, observe $r_t$ and $s_{t + 1}$\;
    $G_t \leftarrow G_{\varnothing}$\;
    Store $s_t$, $a_t$, $r_t$, $s_{t + 1}, G_t$ in $\mathcal{B}$\;
    \If{the episode is over}{
     $G_t \leftarrow \sum_{t' = t}^{T} \gamma^{t' - t} r_{t'}$ for $t \in [0, T]$\;
    }
  }
  \tcc{Update the agent weights (off-policy).}
  \For{each training step}{
    Sample a minibatch $\mathcal{D}_{batch}$ from $\mathcal{B}$\;
    \tcc{Loss expression in Equation~\ref{eqn:loss}.}
    $\theta \leftarrow \theta - \eta \nabla_\theta \mathcal{L}_{SAIL}(\mathcal{D}_{batch})$\;
  }
 }
 \label{alg:1}
\end{algorithm}


\section{Experiments}



In this section, we aim to answer the following questions: \textbf{1)}~How does SAIL perform on various tasks, including hard exploration, using various off-policy algorithms as baselines? (see Sec.~\ref{subsection:sail-dqn} \&~\ref{subsection:sail-iqn}); \textbf{2)}~Does SAIL compare favorably to exploration bonuses? (see Sec.~\ref{subsection:exploration}); \textbf{3)}~What is the impact of stochasticity on SAIL? (see Sec.~\ref{subsection:stochasticity}); and \textbf{4)}
~How does SAIL compare to a more straightforward self-imitation algorithm? (see Sec.~\ref{subsection:sil-vs-sail}).

\subsection{Experimental setting}

\begin{figure}
    \centering
    \vspace{-0.5cm}
    \includegraphics[width=\linewidth]{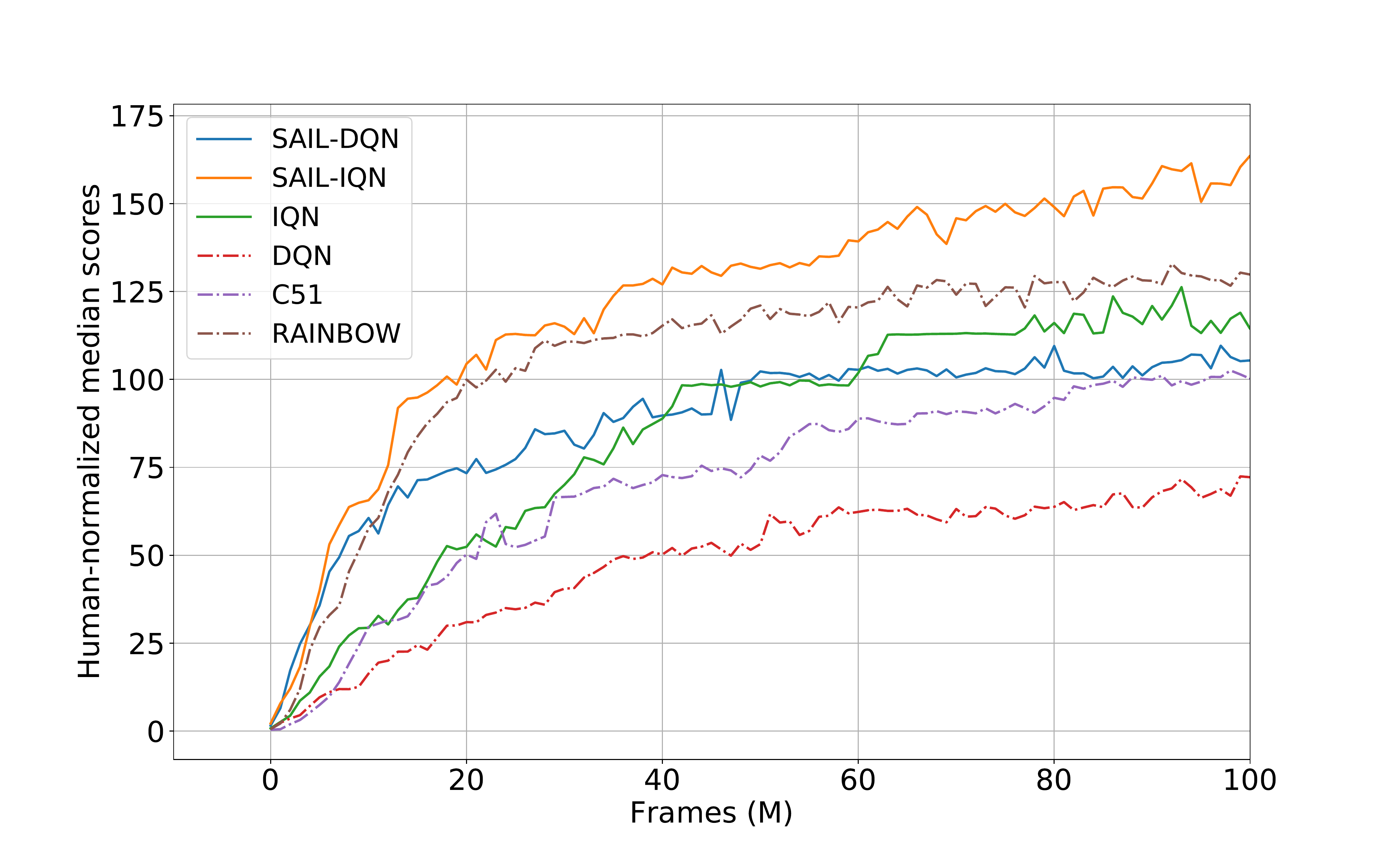}
    \caption{Human-normalized median performance of several off-policy methods. SAIL-IQN outperforms every other method, including Rainbow, which uses prioritized experience replay and n-step bootstrapping. SAIL-DQN, though based on DQN, outperforms C51, a distributional RL algorithm that uses a more advanced optimizer (Adam).}
    \label{fig:11}
\end{figure}

We benchmark our method and baselines on the Arcade Learning Environment~\citep{bellemare2013}, with sticky actions and no episode termination on life losses, following~\citep{machado2018}. Sticky actions make the Atari games stochastic by repeating the agent's previous action with a fixed probability ($0.25$ by default). Hard exploration games are games where local exploration methods fail to achieve high scores. We use the list of hard exploration games from~\citep{bellemare2016}, and set them apart with a bold font in bar plots.
We use the Dopamine framework~\citep{castro2018} to get reference implementations for the agents and setup our experiments. Whether using the base off-policy algorithms as is or in combination with SAIL, we use Dopamine reference hyperparameters for all methods, unless explicitly mentioned. 
%
%
The only SAIL-specific hyperparameter is $\alpha$, that we set to $\alpha=0.9$ in all experiments. The bonus term is also clipped in $[-1,1]$ (as are the rewards in ALE).
We use 6 random seeds for comparative studies on subsets of games, and 3 random seeds for experiments on the 59\footnote{We didn't report results for ElevatorAction, whose ROM failed to load in our setup.} Atari games, due to the computational demand of such experiments.

For bar plots (e.g. Fig.~\ref{fig:1}), the mean relative improvement metric for a method $X$ compared to the baseline $X_{base}$ is, for one game: 
$$(\frac{1}{N} \sum_{t=1}^{200} s_{X, t} - s_{X_{base}, t} )/(| \frac{1}{N} \sum_{t=1}^{200} s_{X_{base}, t} | + \epsilon),$$
where $\{ s_{X, t} \}_{t = 1, \ldots, 200}$ are the game scores for method $X$, collected every million steps in the environment, and averaged across all random seeds.
This metric roughly measures by how much the performance of the proposed method increased or decreased compared to the baseline. 
For aggregated plots (e.g. Fig.~\ref{fig:11}), we report the human-normalized median score, a common metric to evaluate RL methods based on their performance \textbf{across all games}:
$$ (\bar{s}_{X, t} - s_{random})/| s_{human} - s_{random} |,$$
where $\{ \bar{s}_{X, t} \}_{t = 1, \ldots, 200}$ are the median scores across all games for method $X$, collected every million steps in the environment, and averaged across all random seeds. $s_{random}$ is the end score of a random policy, and $s_{human}$ that of the average human player, as reported in~\citep{mnih2013}, and completed for missing games as in~\citep{vieillard2020}. 

\subsection{SAIL-DQN experiments}
\label{subsection:sail-dqn}

\begin{figure}
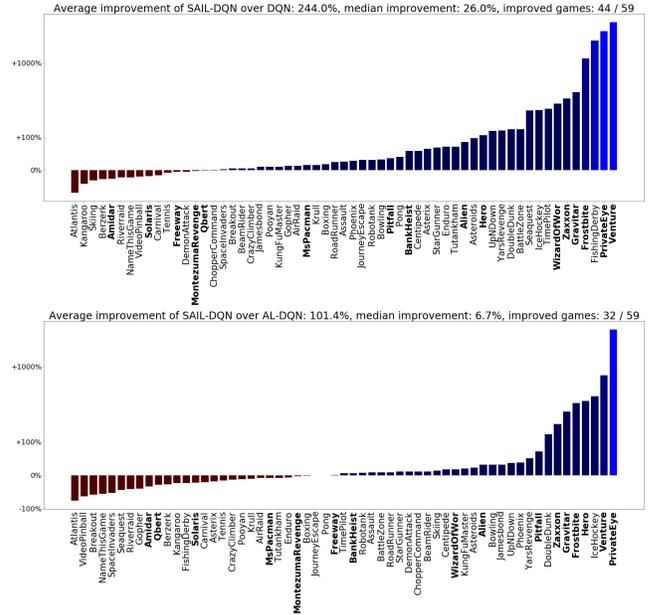

    \centering
    \begin{subfigure}{1\linewidth}
        \includegraphics[width=1\linewidth]{imgs/sil_dqn_vs_dqn_atari.pdf}
    \end{subfigure}
    \begin{subfigure}{1\linewidth}
        \includegraphics[width=1\linewidth]{imgs/sil_dqn_vs_al_dqn_atari.pdf}
    \end{subfigure}
    \caption{SAIL-DQN outperforms both DQN (top) and AL-DQN (bottom), as shown on a relative scale on all Atari games. On hard exploration games (in bold), SAIL-DQN provides a +654.3\% average and +71.8\% median relative improvement over DQN, and a +361.6\% average and +24.9\% median relative improvement over AL-DQN.}
    \label{fig:1}
\end{figure}

We report the performance of SAIL combined to DQN (abbreviated SAIL-DQN) against DQN on all Atari games in Fig.~\ref{fig:1}. We use the Dopamine DQN hyperparameters. 

\begin{figure}
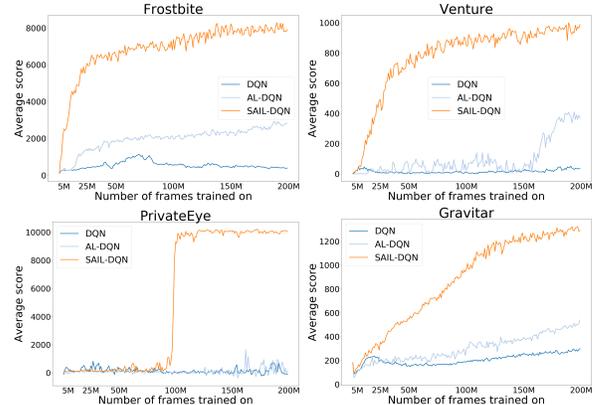

    \centering
    \includegraphics[width=0.45\linewidth]{imgs/sil_dqn_vs_others_Frostbite.pdf}
    \includegraphics[width=0.45\linewidth]{imgs/sil_dqn_vs_others_Venture.pdf}
    \includegraphics[width=0.45\linewidth]{imgs/sil_dqn_vs_others_PrivateEye.pdf}
    \includegraphics[width=0.45\linewidth]{imgs/sil_dqn_vs_others_Gravitar.pdf}
    \caption{Performances of DQN, AL-DQN and SAIL-DQN on four hard exploration Atari games. On PrivateEye, despite sparse rewards, SAIL-DQN outperforms both C51 and IQN.}
    \label{fig:5}
\end{figure}


We also report the same graph with AL-DQN as the baseline on Fig.~\ref{fig:1}. AL-DQN corresponds to the limit case of SAIL where the action-value estimate is always greater than the observed return, and makes for a stronger baseline than vanilla DQN. With the simple modification provided by SAIL, we get a nice average and median performance increase across all games (+101.4\% average and +6.7\% median), the gap being more apparent on hard exploration games where self-imitation shines the most (+361.6\% average and +24.9\% median relative improvement over AL-DQN).

\begin{figure}[t]
    \centering
    \includegraphics[width=\linewidth]{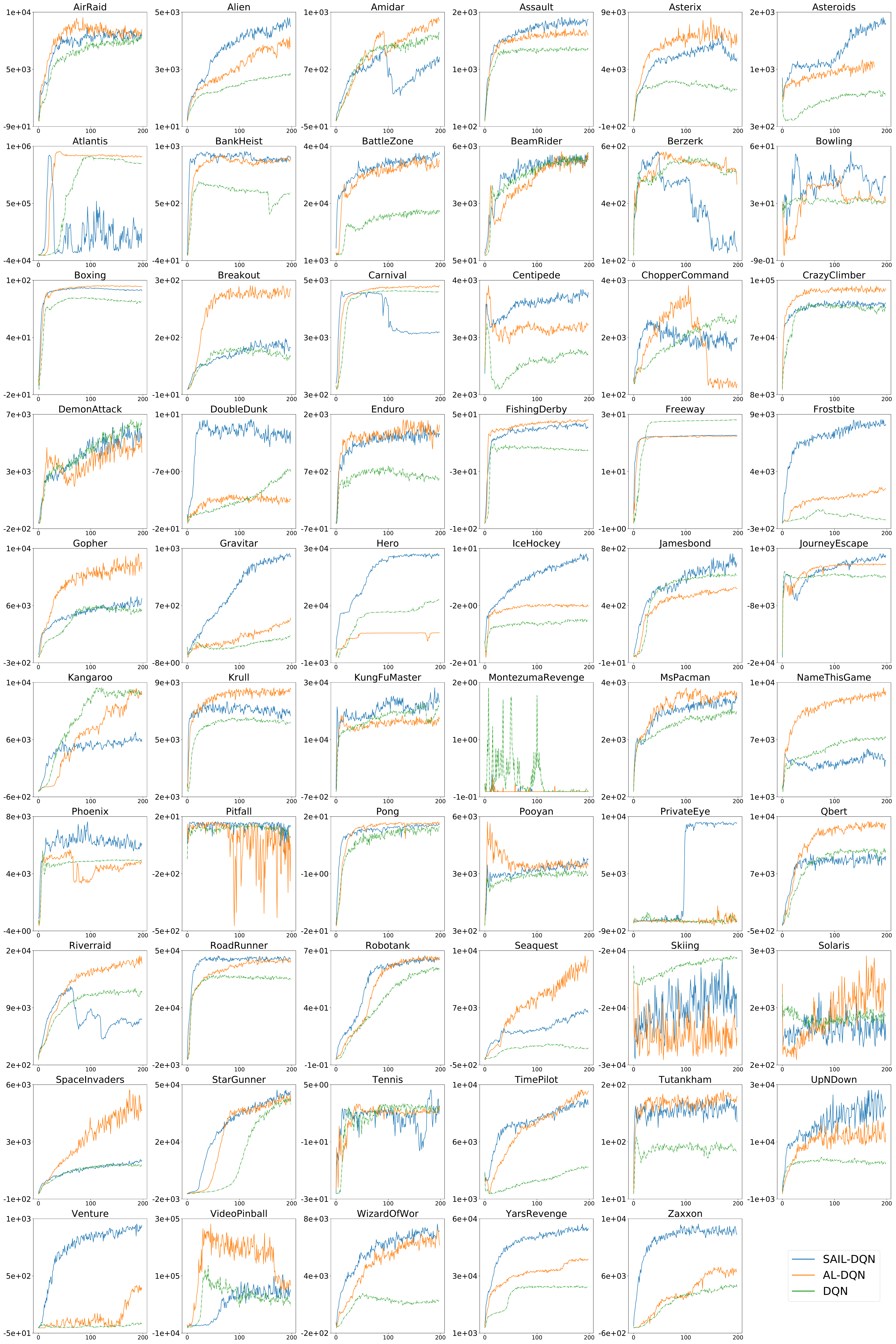}
    \caption{Curves on all games for DQN-based methods.}
    \label{fig:dqn_all}
\end{figure}

We also zoom in on the performances of the three methods (DQN, AL-DQN and SAIL-DQN) on individual games. We choose four hard exploration games: Frostbite, Venture, Gravitar and PrivateEye. The scores are displayed on Fig.~\ref{fig:5}. On Frostbite and PrivateEye, SAIL-DQN outperforms C51\footnote{C51 scores 4250 on Frostbite and 4000 on PrivateEye. Scores are available \href{https://google.github.io/dopamine/baselines/plots.html}{here}.}~\citep{bellemare2017}, a distributional RL agent that uses Adam~\citep{kingma2014}, a more advanced optimizer than DQN's RMSProp~\citep{tieleman2012}. On PrivateEye, that has particularly sparse rewards, SAIL-DQN also outperforms IQN. Finally, we report the final performances on hard exploration games for all three methods in Table~\ref{tab:scores-dqn}, and individual curves on all 59 games can be found in Fig.~\ref{fig:dqn_all}.

\begin{table}
  \caption{Final scores of several off-policy algorithms based on DQN, in hard exploration games.}
  \label{tab:scores-dqn}
  \begin{tabular}{cccc}\toprule
    & \textit{DQN} & \textit{AL-DQN} & \textit{SAIL-DQN} \\ \midrule
    Alien & 2586 & 3514 & \textbf{4769} \\
    Amidar & 1167 & \textbf{1380} & 879 \\
    BankHeist & 594 & 909 & \textbf{915} \\
    Freeway & \textbf{25} & 22 & 23 \\
    Frostbite & 392 & 2833 & \textbf{7912} \\
    Gravitar & 299 & 540 & \textbf{1292} \\
    Hero & 17056 & 7672 & \textbf{30382} \\
    Montezuma's Revenge & 0 & 0 & 0 \\
    MsPacman & 3405 & \textbf{3973} & 3879 \\
    Pitfall! & -84 & -100 & \textbf{-27} \\
    PrivateEye & -114 & 153 & \textbf{10068} \\
    Qbert & 10178 & \textbf{13426} & 9310 \\
    Solaris & 1437 & \textbf{2216} & 855 \\
    Venture & 35 & 371 & \textbf{989} \\
    WizardOfWor & 2020 & 5930 & \textbf{6906} \\
    Zaxxon & 4713 & 6221 & \textbf{10364} \\ \bottomrule
  \end{tabular}
\end{table}

\subsection{SAIL-IQN experiments}
\label{subsection:sail-iqn}

\begin{figure}
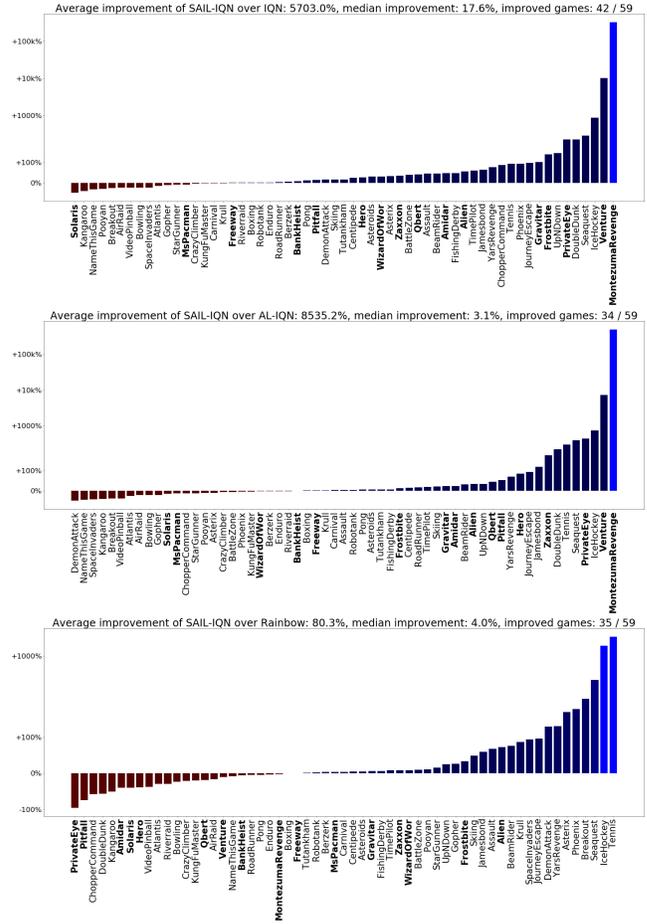

    \centering
    \begin{subfigure}{1\linewidth}
        \includegraphics[width=1\linewidth]{imgs/sil_iqn_vs_iqn_atari.pdf}
    \end{subfigure}
    \begin{subfigure}{1\linewidth}
        \includegraphics[width=1\linewidth]{imgs/sil_iqn_vs_al_iqn_atari.pdf}
    \end{subfigure}
    \begin{subfigure}{1\linewidth}
        \includegraphics[width=1\linewidth]{imgs/sil_iqn_vs_rainbow_atari.pdf}
    \end{subfigure}
    \caption{SAIL-IQN outperforms both IQN (top), AL-IQN (middle) and Rainbow (bottom), as shown on a relative scale on all Atari games. SAIL is responsible for important gains over IQN and AL-IQN on sparse reward hard exploration games such as Montezuma's Revenge, PrivateEye, or Gravitar.}
    \label{fig:6}
\end{figure}

Since SAIL is a general self-imitation method for off-policy learning, it is straightforward to apply it to other off-policy agents.
Hence, having demonstrated strong performance using DQN as a base agent in the previous subsection, we now turn towards a more advanced off-policy algorithm. We choose IQN, a strong distributional RL agent, and use the same IQN hyperparameters as~\cite{dabney2018}, which in contrast to those of Dopamine do not use n-step bootstrapping. 
We report the performance of SAIL combined to IQN (abbreviated SAIL-IQN) against IQN on all Atari games in Fig.~\ref{fig:6}, using the same metric as in previous graphs. We also compare SAIL-IQN to AL-IQN and to Rainbow across all games in the same figure. Despite the fact that Rainbow uses prioritized experience replay and n-step bootstrapping (with $n=3$), we report +80.3\% average and +4\% median relative performance increase for SAIL-IQN over Rainbow.

\begin{figure}
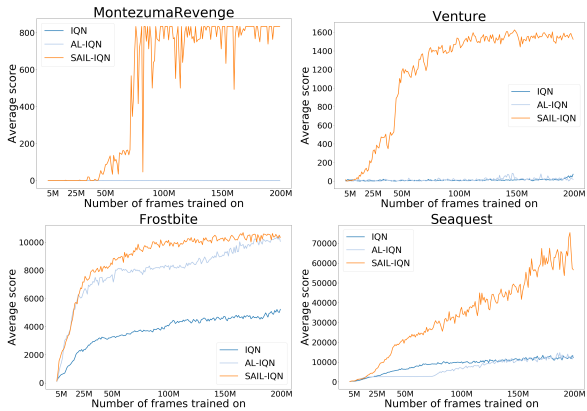

    \centering
    \includegraphics[width=0.45\linewidth]{imgs/sil_iqn_vs_others_MontezumaRevenge.pdf}
    \includegraphics[width=0.45\linewidth]{imgs/sil_iqn_vs_others_Venture.pdf}
    \includegraphics[width=0.45\linewidth]{imgs/sil_iqn_vs_others_Frostbite.pdf}
    \includegraphics[width=0.45\linewidth]{imgs/sil_iqn_vs_others_Seaquest.pdf}
    \caption{Performances of IQN, AL-IQN and SAIL-IQN on three hard exploration Atari games and one easy exploration game (Seaquest), illustrating the versatility of SAIL.}
    \label{fig:indiv-scores-iqn}
\end{figure}

\begin{figure}[t]
    \centering
    \includegraphics[width=\linewidth]{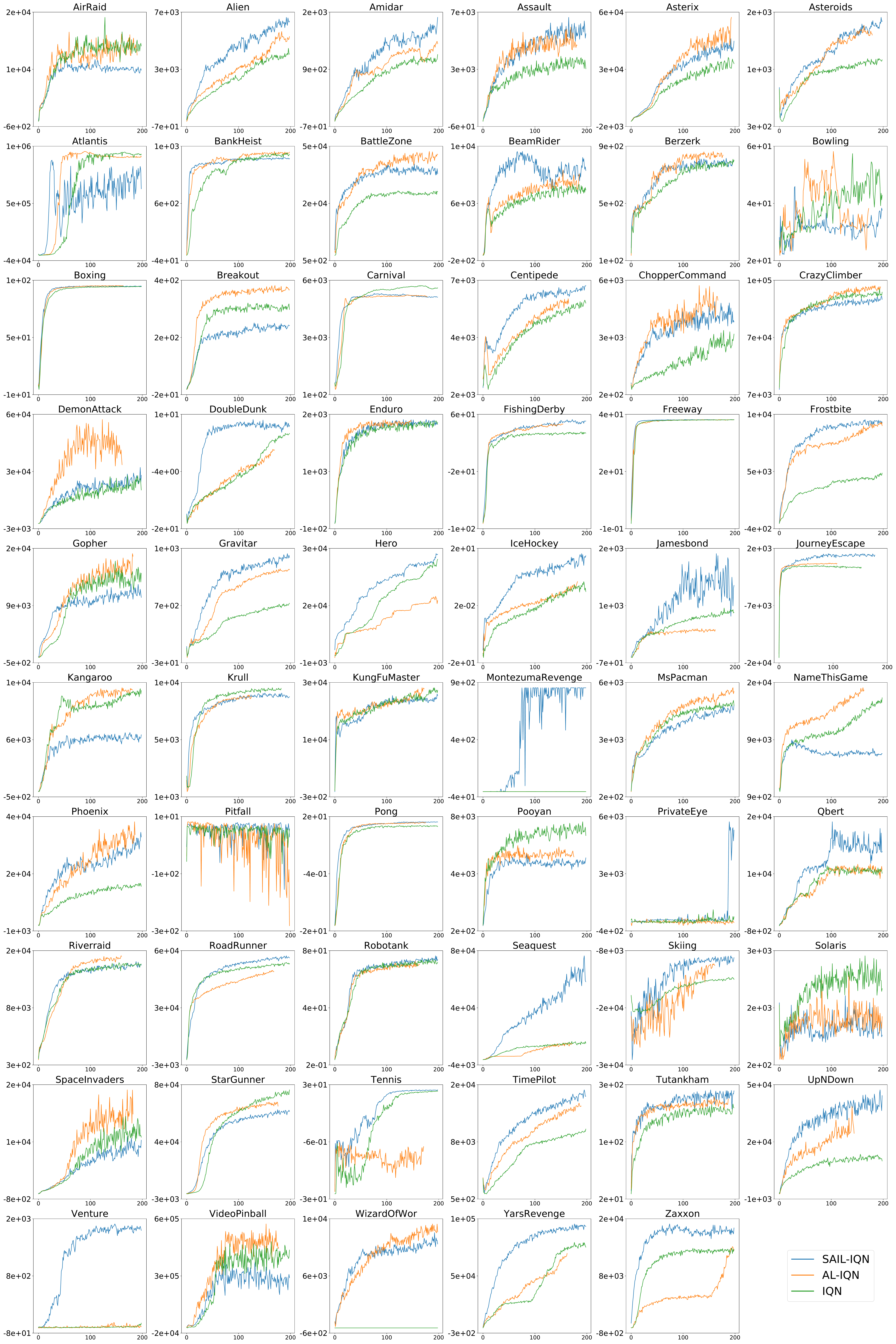}
    \caption{Curves on all games for IQN-based methods.}
    \label{fig:iqn_all}
\end{figure}

Additionally, we display individual game performances of IQN, AL-IQN and SAIL-IQN in Fig.~\ref{fig:indiv-scores-iqn}. We choose two hard exploration games: Montezuma's Revenge and Venture; and also on an easy exploration game: Seaquest, illustrating the versatility of SAIL. We report the final performances on hard exploration games of all three methods in Table~\ref{tab:scores-iqn}, and individual curves on all 59 games in Fig.~\ref{fig:iqn_all}.

\begin{table}[h]
  \caption{Final scores of several off-policy algorithms based on IQN, in hard exploration games.}
  \label{tab:scores-iqn}
  \begin{tabular}{cccc}\toprule
    & \textit{IQN} & \textit{AL-IQN} & \textit{SAIL-IQN} \\ \midrule
    Alien & 4262 & 5318 & \textbf{6245} \\
    Amidar & 1131 & 1311 & \textbf{1752} \\
    BankHeist & 1081 & \textbf{1105} & 1037 \\
    Freeway & 34 & 34 & 34 \\
    Frostbite & 5231 & 10058 & \textbf{10345} \\
    Gravitar & 766 & 1213 & \textbf{1375} \\
    Hero & 28636 & 16804 & \textbf{30083} \\
    Montezuma's Revenge & 0 & 0 & \textbf{833} \\
    MsPacman & 4656 & \textbf{5340} & 4669 \\
    Pitfall! & \textbf{-19} & -280 & -42 \\
    PrivateEye & 357 & 49 & \textbf{4956} \\
    Qbert & 10802 & 11440 & \textbf{15189} \\
    Solaris & \textbf{2165} & 1461 & 992 \\
    Venture & 74 & 19 & \textbf{1525} \\
    WizardOfWor & 5700 & \textbf{10222} & 8933 \\
    Zaxxon & 12024 & 11843 & \textbf{15296} \\ \bottomrule
  \end{tabular}
\end{table}

Finally, we compare several of the algorithms mentioned using the human-normalized median scores in Fig.~\ref{fig:11}, aggregating scores from all games. SAIL-IQN outperforms Rainbow (itself outperforming IQN), without using prioritization or n-step bootstrapping. SAIL-DQN outperforms C51 (itself outperforming DQN).

\subsection{Comparison to Intrinsic Motivation}
\label{subsection:exploration}

\begin{figure}
    \centering
    \includegraphics[width=\linewidth]{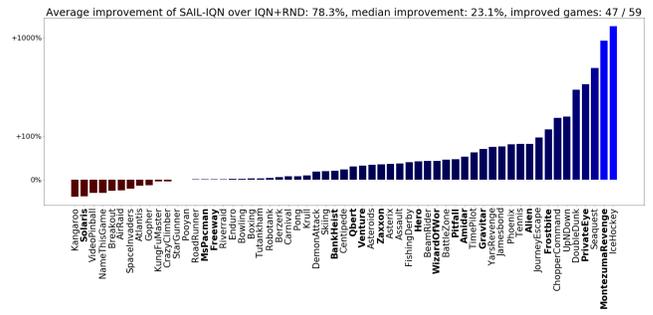}
    \caption{SAIL-IQN outperforms RND-IQN, as shown on a relative scale on all Atari games, including on Montezuma's Revenge, an infamously hard exploration game.}
    \label{fig:12}
\end{figure}

We compare SAIL-IQN to IQN with RND~\citep{burda2018}, a popular intrinsic motivation method for exploration. While the two methods have different motivations, both modify the reward function and were shown to help in hard exploration tasks. RND uses two identical neural networks, the first of which is frozen after initialization. The second network has to predict the output of this random network. The prediction error is then used as a proxy of actual state-visitation counts to reward the agent. For RND, we use code and hyperparameters from~\citet{taiga2019}. RND hyperparameters were calibrated for Rainbow, thus to be fair we use the same RND hyperparameters to study the combination of SAIL and RND (see below), and keep equal SAIL hyperparameters across agents. 
SAIL-IQN shows positive average and median relative improvement scores compared to RND-IQN (+72.2\% average and +28.2\% median relative improvement over RND-IQN, see Fig.~\ref{fig:12}).

\begin{figure}
    \centering
    \includegraphics[width=0.7\linewidth]{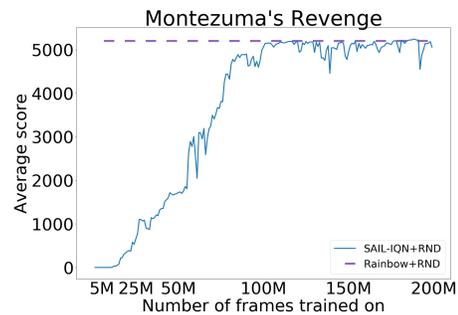}
    \caption{Combining SAIL and RND elevates the performance of IQN and matches that of RND-Rainbow, the best reported in~\citet{taiga2019} on Montezuma's Revenge.}
    \label{fig:montezuma}
\end{figure}

On Montezuma's Revenge, which is an infamously hard exploration game, SAIL-IQN reaches a final score of 833 (for an average score of 513), while RND-IQN reaches a final score of 161 (for an average score of 45), and IQN scores 0 (final and average). SAIL can be combined to RND for further improvements: using RND (as is) and n-step bootstrapping ($n=3$), SAIL-IQN reaches a final score of 5180 on Montezuma's Revenge. This ties SAIL-IQN with RND-Rainbow, which has the strongest score reported in~\citet{taiga2019}. We display the evolution of that score in Fig.~\ref{fig:montezuma}.

While we established the merits of SAIL compared to RND in terms of in-game Atari performance, SAIL has several other advantages over RND. \textit{1) Consistency:} while RND provides with better performance gains on specific games such as Montezuma's Revenge, we find that the gains are not consistent across all hard exploration games, matching the findings of~\citet{taiga2019}. More precisely (not shown), using RND with IQN brought a great performance boost on two games (Montezuma's Revenge and Venture), but did not increase the median performance of IQN (-0.2\% median relative improvement score). \textit{2) Integration:} SAIL requires two simple modifications to off-policy algorithms, saving returns to the replay buffer and modifying the action-value update (1 hyperparameter). In comparison, RND requires two additional networks, separate optimizers and a reward modification (many hyperparameters). We acknowledge that this comparative benefit might be framework-dependent. \textit{3) Compute:} on the same hardware, we find that standard IQN processes an average of 100 frames per second, against 95 for SAIL-IQN and 45 for RND-IQN, so completing experiments is more than twice faster when choosing SAIL over RND.

\subsection{Impact of stochasticity}
\label{subsection:stochasticity}

\begin{figure}
    \centering
    \includegraphics[width=\linewidth]{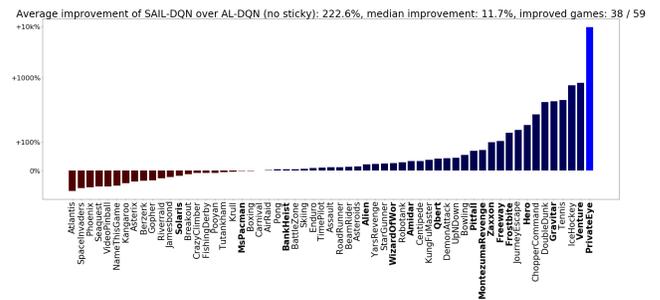}
    \caption{Improvement of SAIL-DQN over AL-DQN on all Atari games, without sticky actions. Result do not diverge from those in the standard ALE setting (+222.6\% versus +101.4\% average, +11.7\% versus +6.7\% median).}
    \label{fig:13}
\end{figure}

To quantify the impact of stochasticity on self-imitation, we perform an ablation and deactivate the sticky actions. Sticky actions~\citep{bellemare2013} introduce stochasticity in otherwise near-deterministic Atari games, by repeating the previous action of the agent with a fixed repeat probability. As in the standard settings of ALE, by default we use a repeat probability of $0.25$, that is here set to $0$. We report results on all Atari games in Fig.~\ref{fig:13}, similarly to the precedent section.

As expected, SAIL brings a larger performance increase when we fall back to a near-deterministic setting but, interestingly, the results do not diverge much from the ones reported with the sticky actions (+231.6\% versus +101.4\% average, +12.7\% versus +6.7\% median).

\begin{figure}
    \centering
    \vspace{-0.4cm}
    \includegraphics[width=\linewidth]{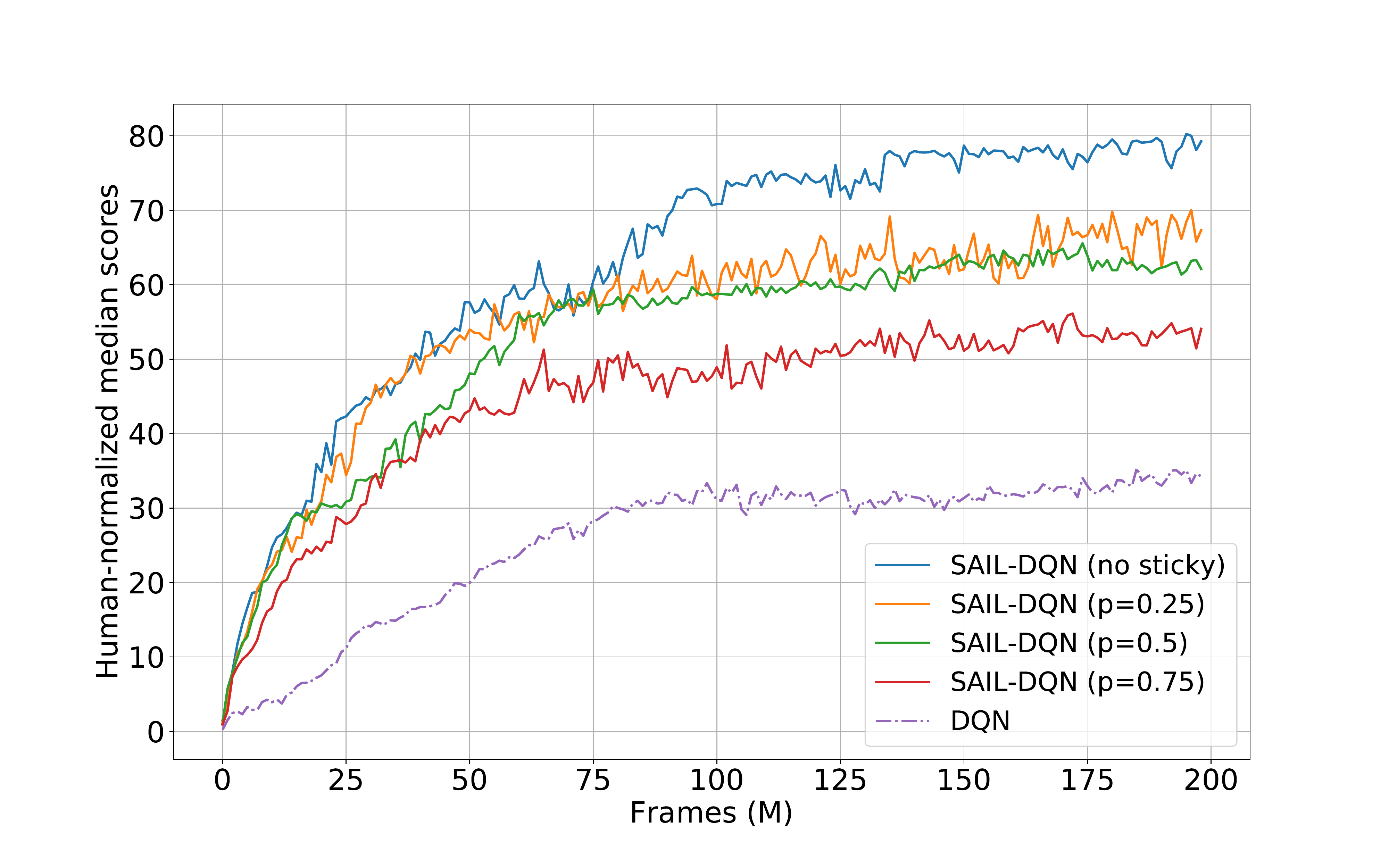}
    \caption{Even under increased stochasticity, SAIL-DQN outperforms DQN in the standard setting ($p=0.25$).}
    \label{fig:stochasticity}
\end{figure}

We extend this experiment and instead increase the repeat probability, making environments more stochastic. We restrict ourselves to hard exploration games to limit the computational cost of the study. As Fig.~\ref{fig:stochasticity} shows, the human-normalized median performance of SAIL-DQN decreases under higher action repeat probabilities. This phenomenon warrants further investigation, as we would have to tell apart the impact of stochasticity on SAIL from the general performance loss due to making the RL problem more difficult, which we leave for future work. Note that the performance of SAIL stays superior to that of DQN in the standard setting.

\subsection{Comparison to straightforward SIL}
\label{subsection:sil-vs-sail}

\begin{figure}
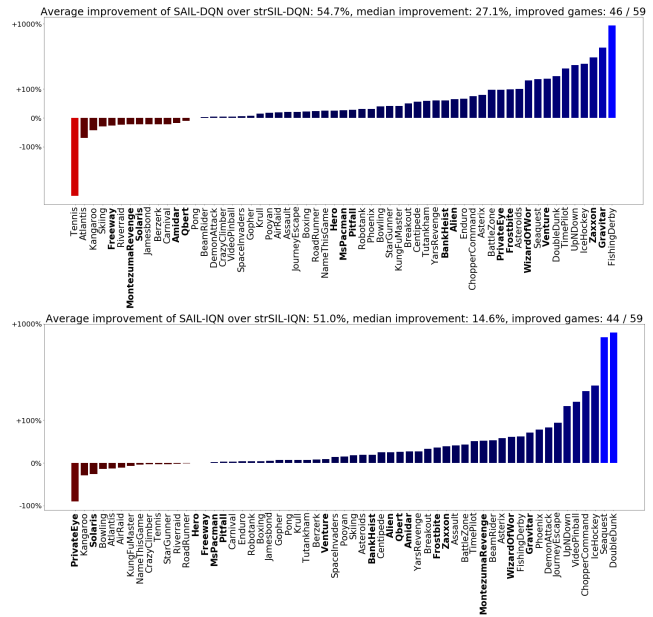

    \centering
    \begin{subfigure}{1\linewidth}
        \includegraphics[width=1\linewidth]{imgs/sil_dqn_vs_straightforward_atari.pdf}
    \end{subfigure}
    \begin{subfigure}{1\linewidth}
        \includegraphics[width=1\linewidth]{imgs/sil_iqn_vs_straightforward_atari.pdf}
    \end{subfigure}
    \caption{Relative improvement of SAIL over  self-imitation (strSIL), based on both DQN (top) and IQN (bottom). On hard exploration games (in bold), SAIL-DQN provides a +72.2\% average and +45.2\% median relative improvement over strSIL-DQN, and SAIL-IQN a +16.0\% average and +22.3\% median relative improvement over strSIL-IQN.}
    \label{fig:straightforward}
\end{figure}

We compare SAIL to the more straightforward self-imitation algorithm depicted Eq.~\eqref{eqn:sil-loss}, the alternate version of SAIL where we use the standard return alone, and a ReLU operator instead of the identity. We display average and median relative improvement scores with both DQN and IQN in Fig.~\ref{fig:straightforward}. 

\section{Related work}


\paragraph{Extending self-imitation learning} 
\citet{guo2018} revisit GAIL~\citep{ho2016}, an adversarial imitation method that encourages the agent to trick a discriminator into taking its behavior for expert behavior. Their method uses the agent's past behavior as expert behavior, identifying promising behavior in a similar way to standard self-imitation. \citet{guo2019} use self-imitation over a diverse set of trajectories from the agent’s past experience, showing that it helps on games where there are local minima that hinder learning. \citet{tang2020} studies the impact of importance sampling corrections and using n-step bootstrapping to replace the observed return in a generalized form of self-imitation, which is studied under the operator view. None of these methods explicitly target off-policy learning algorithms, and as such none have straightforward extensions to that setting.

\paragraph{RL for hard exploration tasks}
Being able to solve hard exploration problems is an important target for RL. So far, quite different strategies have been developed to tackle these problems. Intrinsic motivation methods provide an additional source of reward to the agent, either for going to seldom visited areas of the state space~\citep{bellemare2016, tang2016}, or for experiencing novel things, that challenge its own predictions~\citep{pathak2017, burda2018, savinov2018}. \citet{aytar2018} use expert demonstrations from human player-made videos as imitation data and create auxiliary tasks that incentive close-to-expert progression in the environment. \citet{ecoffet2019} propose an exhaustive exploration method that encourages the agent to visit promising areas of the state space while jointly learning how to get back to those areas if further exploration is needed. \citet{badia2020} combine intrinsic motivation with an episodic motivation based on episodic memory. \citet{paine2019} uses expert demonstrations combined with a recurrent learner to solve partially observable hard exploration tasks. \citet{badia2020agent57} uses a neural network to encapsulate multiple policies with different degrees of exploration and switches between policies using a multi-arm bandit algorithm. While SAIL proves to be useful for hard exploration, its main motivation is to learn properly from the agent's own instructive experiences. In that regard, it is not an exploration method per se, but could generally be combined to said exploration methods for further improvements, as we have illustrated with RND. 

\paragraph{Sparse reward RL}
There is a lot of ongoing effort regarding RL for sparse rewards as well, which is related to hard exploration and overlaps on many environments. In the goal-oriented setting, \citet{andrychowicz2017} propose to modify the replay buffer in hindsight and change the desired goal state by a state the agent actually visited, providing it with free reward signal. \citet{lee2018} accelerate the propagation of sparse reward signals by sampling whole episodes from the replay buffer and performing updates starting from the end of the sampled episode. In a related way, \citet{hansen2018} combine standard off-policy with episodic control, estimating the action-value as the convex combination of the action-value learned off-policy and another estimate from episodic memory. \citet{trott2019} use reward shaping to explore while avoiding local minima. They use sibling trajectories, pairs of trajectories that are different but equally initialized, to create goal states. Since the additions of SAIL to off-policy agents are purely restricted to the action-value updates, SAIL and sparse reward RL techniques could synergize well, which we leave for future experiments.

\paragraph{Misc}
Self-imitation strives to reproduce unexpectedly good behavior, and promotes a form of optimism by doing so, since it does not question the likelihood of the high payoff observed. \citet{he2016} instead show that there are tractable bounds to the optimal action-value function and use them to increase the Q-function when it is inferior to the lower bound (resp. decrease when superior to the upper bound). 
\citet{riquelme2019} propose an algorithmic scheme that combines TD and Monte Carlo estimation in an adaptive way. In contrast, our method uses the observed return for self-imitation, which is the Monte Carlo estimator for the return, while still conserving temporal difference for the action-value update.

\section{Discussion}

In this work, we presented SAIL, a novel self-imitation method for off-policy learning. In comparison with standard self-imitation, SAIL revolves around the action-value function, which makes it compatible with many off-policy algorithms. SAIL is a simple, general and lightweight method that can be equivalently viewed as a modified Bellman optimality operator (that is related to Advantage Learning) or as a modified reward function. Notably, it combines the advantages of self-imitation, that is better reinforcement in sparse reward scenarios, and those of Advantage Learning, that is increasing the action-gap. We studied its performance in the Arcade Learning Environment and demonstrated that SAIL brought consistent performance gains, especially on hard exploration games, at virtually no cost.

\newpage

\bibliographystyle{ACM-Reference-Format}
\bibliography{biblio}

\end{document}